\documentclass[sn-vancouver,Numbered]{sn-jnl}

\usepackage{graphicx}%
\usepackage{multirow}%
\usepackage{amsmath,amssymb,amsfonts}%
\usepackage{amsthm}%
\usepackage{mathrsfs}%
\usepackage[title]{appendix}%
\usepackage{xcolor}%
\usepackage{textcomp}%
\usepackage{manyfoot}%
\usepackage{booktabs}%
\usepackage{algorithm}%
\usepackage{algorithmicx}%
\usepackage{algpseudocode}%
\usepackage{listings}%
\usepackage{natbib}
\usepackage{tabularx}
\usepackage{lmodern}

\begin{document}

\title[Article Title]{\texttt{shap-select}: Lightweight Feature Selection Using SHAP Values and Regression}

\author*[1]{\fnm{Egor} \sur{Kraev}}\email{egor.kraev@wise.com}

\author[1]{\fnm{Baran} \sur{Koseoglu}}\email{baran.koseoglu@wise.com}

\author[1]{\fnm{Luca} \sur{Traverso}}\email{luca.traverso@wise.com}

\author[1]{\fnm{Mohammed} \sur{Topiwalla}}\email{mohammed.topiwalla@wise.com}

\affil[1]{\orgdiv{Data Science}, \orgname{Wise Plc}, \orgaddress{\street{Shoreditch High St}, \city{London}, \postcode{E1 6JJ}, \country{UK}}}

\abstract{Feature selection is an essential process in machine learning, especially when dealing with high-dimensional datasets. It helps reduce the complexity of machine learning models, improve performance, mitigate overfitting, and decrease computation time. This paper presents a novel feature selection framework, \texttt{shap-select}. The framework conducts a linear or logistic regression of the target on the Shapley values of the features, on the validation set, and uses the signs and significance levels of the regression coefficients to implement an efficient heuristic for feature selection in tabular regression and classification tasks. We evaluate \texttt{shap-select} on the Kaggle credit card fraud dataset, demonstrating its effectiveness compared to established methods such as Recursive Feature Elimination (RFE), HISEL (a mutual information-based feature selection method), Boruta and a simpler Shapley value-based method. Our findings show that \texttt{shap-select} combines interpretability, computational efficiency, and performance, offering a robust solution for feature selection.}


\maketitle

\section{Introduction}\label{sec1}

Feature selection is a fundamental step in the machine learning pipeline that aims to select the most relevant features from the data, discarding those that are irrelevant or redundant. It plays a crucial role in improving the performance of machine learning models, reducing overfitting and complexity of the machine learning models, and enhancing interpretability \cite{guyon2003introduction}. Compared to the other dimensionality reduction techniques based on projection, feature selection methodologies do not alter the representations of the original features but selects a subset of them. With the increasing availability of large datasets and ever growing number of features used in training machine learning models, especially in domains like healthcare, finance, and bioinformatics, feature selection has become even more important.

In high-dimensional datasets, irrelevant or redundant features can introduce noise and increase the risk of overfitting. This is particularly problematic in domains such as healthcare, where models are expected to make critical decisions. For instance, in the prediction of patient outcomes, irrelevant features might skew the model's predictions, leading to incorrect diagnoses or treatment plans \cite{saeys2007review}. Similarly, in finance, detecting fraudulent transactions requires accurate models that can distinguish between normal and fraudulent behaviors, often based on subtle patterns. Redundant or irrelevant features can reduce the model's ability to detect such patterns, leading to increased false positives or negatives.

The challenge of feature selection becomes even more pronounced with the development of more complex machine learning models, such as deep neural networks. These models, although powerful, can easily overfit to high-dimensional data unless careful feature selection is applied. In this context, the development of efficient and interpretable feature selection methods has become a pressing need in the field of machine learning.

Training a model for all combinations of features and evaluating becomes an NP-hard problem as the number of features increase with complexity $ O(2^N)$. Therefore it is essential to consider computational efficiency of a feature selection methodology as well as its effectiveness. Feature selection methodologies can broadly be categorized into three main types: filter methods, wrapper methods, and embedded methods. Filter methods assess the importance of features by using statistical tests independent of the model. Wrapper methods evaluate subsets of features by training a model normally, possibly on different feature combinations, and applying post-processing. Finally, embedded methods incorporate feature selection into the model training process, often leveraging regularization techniques like Lasso or decision tree-based algorithms \cite{chandrashekar2014survey}.

In this paper, we introduce \texttt{shap-select}, a feature selection framework integrating SHAP values \cite{lundberg2017unified} with statistical significance testing to provide an efficient heuristic for feature selection. The framework runs a regression of the target on the SHAP values of the original features, on the validation set, and filters features based on their coefficients and statistical significance.  We demonstrate its effectiveness on the Kaggle credit card fraud detection dataset \cite{kaggle}, comparing it with other feature selection methods such as Boruta, Recursive Feature Elimination (RFE), High-dimensional Information-based Selection (HISEL), and a simpler Shapley-value based method implemented in the \texttt{\texttt{shap-select}ion} package. 

\section{Related Work}\label{sec2}

Feature selection has been a topic of extensive research, with various methods proposed in the literature to handle the challenges posed by high-dimensional datasets. The most commonly used feature selection techniques fall into three categories: filter methods, wrapper methods, and embedded methods.

\subsection{Filter Methods}
Filter methods use ranking techniques to rank features based on their relevancy to the dependent variable, without reference to a regression/classification model. Features are ranked according to a scoring function $S(i)$ which calculates a value for each feature $i$ using a set of n samples $(x_{t,i})$ and targets $y_t$ where $t=1,..n$. Features which score above a certain threshold are selected and the model is trained on the selected subset of features. The definition of relevancy of a feature differs from one application to another \cite{kohavi1997wrappers}. Some popular filter methods include Pearson correlation, mutual information, and Chi-square tests. These methods are computationally efficient since they don't require training a machine learning model and they are proven to work well for certain datasets \cite{lazar2012survey}. However, they are often criticized for their inability to account for feature interactions, as they treat each feature independently \cite{brown2012conditional}.

In the context of large datasets, filter methods have been extensively used due to their scalability. For example, Zhao and Liu \cite{zhao2007spectral} proposed a spectral feature selection method based on spectral graph theory, which scales well to large datasets while maintaining computational efficiency. Although these methods are effective in reducing dimensionality, they often lack the ability to consider interactions between features and the target variable. High-dimensional Information-based Selection (HISEL) has also been proposed for handling large datasets, leveraging mutual information criteria to evaluate feature dependencies in high-dimensional contexts. HISEL applies criteria such as max-dependency, max-relevance, and min-redundancy to identify informative features without inducing redundancy, making it useful for large datasets \cite{peng2005feature}. Despite its strengths, HISEL can be computationally intensive, and like many filter methods, may not fully capture feature interactions affecting the target variable.

Another limitation of filter methods is their susceptibility to noise. Because they rely solely on statistical measures, filter methods can inadvertently select irrelevant features that appear important due to random correlations in the data. This is especially problematic in noisy datasets, where statistical significance may not correspond to predictive relevance \cite{peng2005feature}. To mitigate this, researchers have proposed hybrid methods that combine filter techniques with more sophisticated methods, such as embedded or wrapper approaches \cite{huynh2021combining}.

\subsection{Wrapper Methods}
Wrapper methods, introduced by Kohavi and John \cite{kohavi1997wrappers}, evaluate feature subsets by training a model normally, possibly on different feature subsets, and selecting the subset that provides the best performance. One well-known wrapper method is Recursive Feature Elimination (RFE), which iteratively removes the least important features based on model coefficients. While wrapper methods often result in better performance, they usually are computationally expensive since they require repeated training of the model on different feature subsets.

Wrapper methods tend to perform well in scenarios where feature interactions are important, but the typically high computational cost limits their applicability to big datasets. For instance, Guyon et al. \cite{guyon2002gene} successfully used RFE for gene selection in cancer classification tasks. However, as the size of the dataset increases, wrapper methods become less practical due to the iterative trial-and-error process.

A major drawback of wrapper methods is their tendency to overfit on smaller datasets. Since wrapper methods optimize model performance based on training data, they can become overly tuned to the idiosyncrasies of the training set, reducing generalization to unseen data \cite{zhao2010advancing}. Furthermore, the computational cost of many wrapper methods increases exponentially with the number of features, making them impractical for datasets with thousands of features \cite{li2017feature}.

\subsection{Embedded Methods}
Embedded methods incorporate feature selection directly into the model training process. Techniques like Lasso (Least Absolute Shrinkage and Selection Operator) \cite{tibshirani1996regression} use L1 regularization to penalize less important features, effectively shrinking their coefficients to zero. Decision tree-based models, such as Random Forests or XGBoost, inherently rank features by their importance in the model's decision-making process \cite{chen2016xgboost}.

While embedded methods are often more computationally efficient than wrapper methods, they can suffer from overfitting, especially when the model is complex or the dataset is noisy. Boruta \cite{kursa2010boruta}, an embedded method based on Random Forests, addresses this issue by iteratively permuting feature importance values to test their statistical significance.

Embedded methods have the advantage of being integrated directly into the learning algorithm, which allows them to account for feature interactions naturally. However, they are not without their limitations. Embedded methods like Lasso can be sensitive to the choice of regularization parameters, which may lead to suboptimal feature selection if not carefully tuned \cite{zou2005regularization}. Additionally, while decision tree-based methods can handle interactions between features, they are prone to overfitting in the presence of noisy data or small sample sizes \cite{breiman2001random}.

\subsection{Methods using Shapley Values for feature selection}
Many feature selection methods make use of Shapley values. These values provide a measure of the contribution of each feature to a model's prediction. The Shapley value for feature \(i\) is computed as:

\[
\phi_i = \sum_{S \subseteq N \setminus \{i\}} \frac{|S|!(|N|-|S|-1)!}{|N|!} \left[ f(S \cup \{i\}) - f(S) \right]
\]

Where \( S \) is a subset of features, \( N \) is the full set of features, and \( f(S) \) represents the model output for the subset \( S \) \cite{Shapley1952}. Shapley values have the unique property of local accuracy, meaning that the sum of Shapley values for all features is equal to the model's prediction, for each data point. 

While Shapley values can be expensive to compute in the general case, there is an efficient way to compute them for tree-based models, in particular for the popular gradient-boosting models such as XGBoost, CatBoost, and LightGBM, as implemented in the \texttt{shap} package.

Most methods that use Shapley values for feature selection, such as \cite{shay2005, powershap2020, MarcilioJr2020shapselection}, and similar feature selection methods identified in recent literature \cite{shapley2021}, predominantly rank features based on absolute SHAP values without employing statistical significance testing.  The package \texttt{shap-selection}, which we use as one of the comparison approaches, is an implementation of \cite{MarcilioJr2020shapselection}.

The approaches that use Shapley values for feature selection also almost uniformly work in-sample, without considering a validation set (or at most using a validation set to decide when to stop iterative feature removal \cite{vanMourik2023}); and use some kind of bootstrapping to derive significance thresholds to decide which features should be discarded. These are quite computation-intensive as they require multiple fits of the original model. 

The one exception is \cite{Sebastian2024}, which uses the Shapley values to analyze whether each feature, on average, has a positive, negative, or irrelevant influence on closeness of model output to the target, and looks at distinct training, validation, and test sets. However, it also requires multiple fits of the original model and is thus computationally demanding.

\section{Methodology}

\subsection{Motivation}
In developing \texttt{shap-select}, we were inspired by Mazzanti’s work on accelerating feature selection via approximate predictions \cite{mazzanti2020approximate}. However, we went one step further by noticing that, as the Shapley values must add up to the model output, we can now use the statistical apparatus around linear regression, such as coefficient significance values. 

This unlocks a set of methods and intuitions from the linear regression domain, and generalizes to classification problem by replacing linear regression with logistic, for example expressing the feature selection threshold as a statistical significance, making its values easy to interpret. 

Additionally, the framework applies Bonferroni correction \cite{armstrong2014use} to adjust for multiple testing when calculating feature significance in the multiclass classification case. These enhancements improve both the interpretability and robustness of the selection process.

\subsection{Regression models}
In the case of regression (predicting a floating-point target), the procedure is especially simple: we train the original model (for which we're doing feature selection) on the training set. We use that trained model to calculate the Shapley values of all the features on the validation set.  Finally, we run a linear regression of the target on the Shapley values of all the features, on the validation set.  If our model was perfect, all the coefficients would be exactly equal to one, as the model prediction, which equals the sum of all the Shapley values, would equal the target. But in reality, that will of course not be the case.
How can the outcome of that regression tell us whether a particular feature is useful? First of all, if the coefficient is not statistically significant, then it does not systematically bring the model prediction closer to the target, and so should probably be dropped (pending the choice of significance threshold). Another important case is that if the coefficient is significant but negative, then dropping the variable (i.e. setting the coefficient to zero) will on average bring the model prediction closer to the target than leaving it in (ie setting the coefficient to zero). So, all the features with negative coefficients in that regression should just be dropped, regardless of the significance threshold.
\subsection{Binary classifier models}
For classifier models, the logic is very similar. The only difference is that now Shapley values explain the log-odds of the classifier score, and we fit a logistic regression instead of a linear one; but the apparatus of statistical significance is still available, and the interpretation of negative coefficients is the same.
\subsection{Multiclass classifier models}
One challenge here is that we don't just get one Shapley value per datapoint and feature, but a vector, one value per class. As a first step, we train a separate logistic regression for each class, in much the same way we do for binary classifiers. But how do we aggregate these? We want to both get the score signifying the highest significance (if a feature is important for predicting at least one class, it is important), and still discard the negative coefficients, which you can't glean from the significance scores alone. An elegant way of doing that is just taking the largest t-value (across all classes) for each feature, without taking the absolute value; and then convert it to a significance value if it's positive or else discard the feature.
The second challenge is data mining: for example, if we set a 5\% significance threshold and evaluate it on 20 different regressions, on average we'll get one false positive. To avoid this, we apply a Bonferroni correction, by multiplying the significance that we inferred from the t-value by the number of classes (usually it's done the other way around, by adjusting the threshold, but we feel this way makes it easier to apply the same downstream threshold comparison logic for all model types).
\subsection{Dealing with collinearity}
In general, taking individual coefficient significance as the sole guide to the usefulness of a feature is dangerous in the case of features with high collinearity. For example, if we have two identical features with high predictive power, both their coefficients will be individually insignificant, but we only want to drop one of the features from the featureset, not both. To deal with that, we firstly apply a tiny (weight of 1e-6) L1 regularization to all the regressions above. Just in case that might not be enough, we apply the procedure described above in a recursive fashion: once we receive the linear/logistic regression results, with a t-value associated with each feature, we drop the feature with the lowest t-value, and repeat the regression on the remaining Shapley value features, iterating until we've run out of features. This way, in the case of collinearity we will first only discard one of the affected variables, then the next one, etc until the collinearity is eliminated, as illustrated in the following pseudocode:

\vspace{0.5cm}

\begin{verbatim}
function SHAP_SELECT(model, data, target, threshold, alpha):
    shap_values = compute_shap_values(model, data)
    task = infer_task_type(target)
    
    p_values_df = compute_significance(shap_values, target, task, alpha)
    
    while features_remain:
        feature_to_remove = feature_with_lowest_significance(p_values_df)
        shap_values = remove_feature(shap_values, feature_to_remove)
        p_values_df = update_significance(shap_values, target, task, alpha)

    selected_features = select_significant_features(p_values_df, threshold)
    return selected_features
\end{verbatim}

\vspace{0.5cm}

The pseudo-code begins by calculating the SHAP values for each feature in the dataset, which measure the contribution of each feature to the model’s prediction. The task type (binary, multiclass, or regression) is inferred automatically if not specified by the user. Statistical significance is then computed using logistic regression for classification tasks or Ordinary Least Squares (OLS) for regression tasks, based on the SHAP values.

Once the p-values are calculated, the framework enters an iterative process where the least significant feature is removed, and the significance of the remaining features is re-evaluated. This process continues until no more features fall below the defined significance threshold. The final step selects and returns the features that meet the significance criteria, balancing model performance and feature interpretability.

\section{Experiments}\label{sec4}

In this section, we evaluate the performance of \texttt{shap-select} using the Kaggle credit card fraud detection dataset \cite{kaggle}. The dataset contains 284,807 transactions, of which 492 are labeled as fraudulent (a severe class imbalance), and includes 30 features such as time, amount, and principal components derived from a Principal Component Analysis (PCA) transformation. The goal is to detect fraudulent transactions, which is a binary classification task.

\subsection{Experiment Design}

For our experiments, we split the dataset into train 0.60, validation 0.20 and test 0.20 sets. We used the XGBoost classifier across all feature selection methods to ensure consistency in performance comparison and trained the model on train set whereas test set was used to evaluate the performance of the feature selection methods. The XGBoost hyperparameters used for the experiments were:

Objective: binary:logistic
Evaluation Metric: logloss
Number of Boosting Rounds: 100
Verbosity: 0 (silent)
Seed: 42
Nthread: 1
These parameters were fixed across all methods to prevent model training differences from skewing the comparison of feature selection techniques. The dataset was preprocessed without scaling or normalization, as XGBoost can handle raw data.

All experiments were conducted on a machine equipped with an Apple M1 Pro chip and 32 GB of RAM, running macOS, and utilizing Python 3.9. We used the following libraries to conduct the experiments: \texttt{shap-select} \cite{kraev2024shapselect}, \texttt{shap-select}ion \cite{MarcilioJr2020shapselection}, scikit-learn \cite{pedregosa2011scikit}, xgboost \cite{chen2016xgboost}, skfeature-chappers \cite{li2018feature}, and Boruta \cite{kursa2010feature}. We used default parameters with the packages used in the experiments and used 15 as number of features to be selected for \texttt{shap-select}ion and RFE.

\subsection{Methods}

The following feature selection methods were tested:

\textbf{\texttt{shap-select}}: Our proposed method, \texttt{shap-select}, combines SHAP values with statistical significance testing. SHAP values provide a unified approach for interpreting model predictions by offering insights into feature importance \cite{lundberg2017unified, lundberg2018explainable}. Statistical tests are applied to these SHAP values to retain only the most statistically significant features.

\textbf{\texttt{shap-select}ion}: Based on the approach by Marcílio and Eler \cite{MarcilioJr2020shapselection}, this method uses SHAP values to rank feature importance and selects the top features based on absolute SHAP values. This is a model-agnostic approach and provides straightforward interpretability through SHAP explanations.

\textbf{Recursive Feature Elimination (RFE)}: RFE is a wrapper-based method that iteratively removes the least significant features, as indicated by model coefficients, and retrains the model in each step. This is especially effective in capturing feature interactions and has shown strong results in various classification tasks \cite{guyon2002gene, guyon2003introduction}.

\textbf{HISEL (High-dimensional Information-based Selection)}: HISEL, based on Maximum Relevance Minimum Redundancy (MRMR), maximizes dependency between features and the target variable while minimizing redundancy between selected features. This mutual information-based filter method may lack sensitivity to feature interactions \cite{li2018feature}.

\textbf{Boruta}: Boruta is a wrapper method that builds on the Random Forest classifier's feature importance scores. It iteratively compares the importance of each real feature to that of a randomized shadow feature, thereby identifying truly important features with robust statistical support \cite{kursa2010feature}.

Each method was used to select features, followed by training an XGBoost model on the selected features \cite{chen2016xgboost}. Model performance on the test set was evaluated using the following metrics:

\begin{itemize} \item \textbf{Accuracy}: The proportion of correctly classified samples. \item \textbf{F1 Score}: The harmonic mean of precision and recall, particularly useful for our imbalanced dataset. \item \textbf{Runtime}: The time taken to perform feature selection (excluding training time), providing insight into computational efficiency. \end{itemize}

This setup allows for a broad comparison across feature selection methods, balancing computational efficiency, interpretability, and predictive performance.

\subsection{Results}
\subsubsection{Comparison to other methods}
\begin{table}[htbp]
\centering
\begin{tabularx}{\textwidth}{c|c|c|c|c} 
 \hline
\textbf{Method}      & \textbf{Selected Features} & \textbf{Accuracy} & \textbf{F1 Score} & \textbf{Runtime (s)} \\
\hline
\textbf{\texttt{shap-select}}       & \textbf{6}                         & \textbf{0.999596}           & \textbf{0.870056}              & \textbf{21.800304}                 \\
\texttt{shap-selection}                & 15                          & 0.999596           & 0.868571              & 7.052642                 \\
Boruta               & 11                       & 0.999631           & 0.881356	              & 95.853016                \\
RFE                  & 15                         & 0.999561           & 0.857143              & 12.907900                \\
HISEL               & 30                       & 0.999561           & 0.858757	              & 109.028664                \\
No Feature Selection               & 30                       & 0.999561           & 0.858757	              & 1.561542                \\
\hline
\end{tabularx}
\caption{Performance of feature selection methods on credit card fraud dataset.}
\label{tab:results}
\end{table}

The results of the experiments are summarized in Table \ref{tab:results}. \texttt{shap-select} demonstrates competitive performance in terms of accuracy and F1 score while maintaining reasonable runtime compared to other methods. 

\texttt{shap-select} offered the best balance between model performance and computational cost, selecting 6 features with a runtime of 21 seconds. HISEL, although selecting all 30 features, had a higher runtime of 109 seconds and had one of the lowest F1 score. RFE demonstrated faster runtime but with lower F1 score and lower accuracy than \texttt{shap-select}. Boruta performed better than \texttt{shap-select} in terms of accuracy and F1 score, however runtime is much more higher than \texttt{shap-select}. \texttt{shap-select}ion came closest to \texttt{shap-select} performance, but still couldn't quite match it. 

It is noteworthy that \texttt{shap-select} selected the smallest number of features, yet resulted in the best performance on the test set.
\subsubsection{Threshold selection}
\begin{figure}[htbp]
    \centering
    \includegraphics[width=0.7\linewidth]{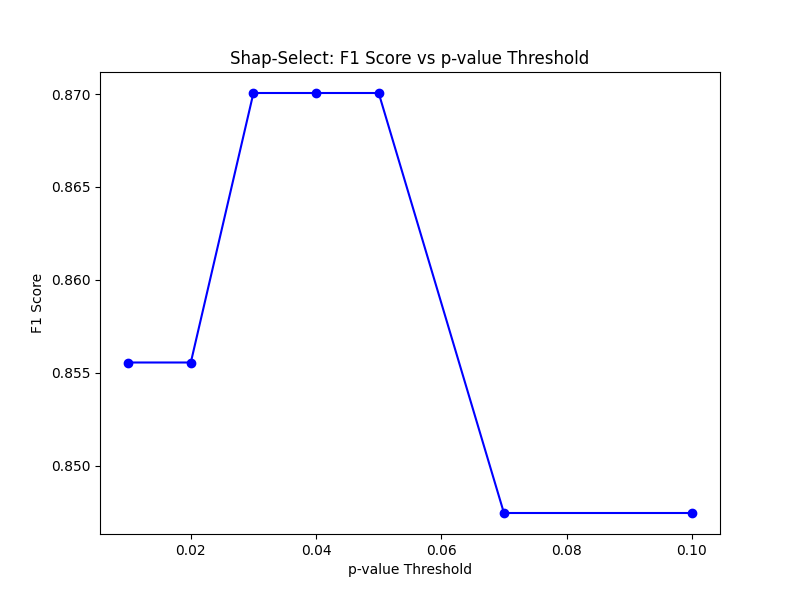} 
    \caption{Model performance as a function of the confidence threshold for feature selection.}
    \label{fig:thresholds}
\end{figure}
By default, \texttt{\texttt{shap-select}} uses the usual 5\% confidence threshold for selecting features. Does this lead to optimal model performance?

Figure~\ref{fig:thresholds} shows the F1 values for running our approach at different thresholds, with a smaller threshold corresponding to fewer features being selected. We see that for this dataset at least, 5\% is a reasonable threshold value.
\subsubsection{Would iteration help?}
As the original model fitted to all the features is not necessarily the best possible way of discovering the potential usefulness of features, it is reasonable to ask whether, after selecting the features with \texttt{\texttt{shap-select}} and fitting the original model on the selected features, we should repeat the procedure using the newly fitted model and the reduced featureset. However, in our tests, when the repeated application of the procedure led to eliminating additional features, that reduced the resulting model performance instead of improving it. So a single pass of \texttt{\texttt{shap-select}} seems to be the optimal way of using it.

\section{Software Availability}

The \texttt{shap-select} library, which implements our feature selection framework described in this paper, is available as an open-source Python package. The codebase can be accessed at \url{https://github.com/transferwise/shap-select}, where users can find the source code, documentation, and examples on how to integrate the framework into their machine learning workflows.

We request users of the \texttt{\texttt{shap-select}} library to cite this paper and to cite following in their publications \cite{kraev2024shapselect}:

Kraev, E., Koseoglu, B., Traverso, L., Topiwalla, M., 2024. \texttt{shap-select}: A SHAP-based Feature Selection Library. Available at: \url{https://github.com/transferwise/shap-select}.

\section{Conclusion}\label{sec5}
In this paper, we presented \texttt{shap-select}, a novel embedded feature selection framework that leverages SHAP values combined with statistical significance testing. We demonstrated the framework's effectiveness using the Kaggle credit card fraud dataset, where it outperformed traditional methods such as HISEL, RFE and \texttt{shap-select}ion in terms of model performance while combining computational efficiency, interpretability, and performance.

\texttt{shap-select} is computationally efficient because it calculates SHAP values once and iteratively removes the least significant feature via linear/logistic regressions, without retraining the model on different subsets. This makes it particularly suitable for large, high-dimensional datasets where reducing the number of features without sacrificing model performance is critical.

\texttt{shap-select} introduces a unique combination of model interpretability, statistical rigor, and computational efficiency, making it a valuable tool for feature selection in high-dimensional datasets. Future work could focus on extending the framework to support more complex models and tasks. Additionally, incorporating other statistical techniques, such as false discovery rate control, could further enhance the framework's robustness in high-dimensional datasets.

\end{document}